\pdfoutput=1

\documentclass[11pt]{article}

\usepackage[]{acl}

\usepackage{times}
\usepackage{latexsym}
\usepackage{booktabs}
\usepackage{multirow}
\usepackage{graphicx}
\usepackage{subcaption}
\usepackage{amsmath}
\usepackage{amsfonts} 
\usepackage{float}
\usepackage{soul}
\usepackage{tikz}
\usepackage{hyperref}

\newcommand{\textttt}[1]{\texttt{\small{#1}}}

\usetikzlibrary{shapes,decorations,arrows,calc,arrows.meta,fit,positioning}
\tikzset{
    -Latex,auto,node distance =1 cm and 1 cm,semithick,
    state/.style ={ellipse, draw, minimum width = 0.7 cm},
    point/.style = {circle, draw, inner sep=0.04cm,fill,node contents={}},
    bidirected/.style={Latex-Latex,dashed},
    el/.style = {inner sep=2pt, align=left, sloped}
}

\usepackage[T1]{fontenc}

\usepackage[utf8]{inputenc}

\usepackage{microtype}

\title{Interventional Probing in High Dimensions: 
An NLI Case Study}

 \author{Julia Rozanova$^{1}$,~ Marco Valentino$^{2}$,~ Lucas Cordeiro$^{1}$,~ Andr\'{e} Freitas$^{1,2}$\\
University of Manchester, United Kingdom$^{1}$ \\
\texttt{\{firstname.lastname\}@manchester.ac.uk}$^{1}$\\
Idiap Research Institute, Switzerland$^{2}$ \\
\texttt{\{firstname.lastname\}@idiap.ch}$^{2}$\\
}

\begin{document}
\maketitle
\begin{abstract}
Probing strategies have been shown to detect the presence of various linguistic features in large language models; in particular, semantic features intermediate to 
the ``natural logic" fragment of the Natural Language Inference task (NLI).
In the case of natural logic, the relation between the intermediate features and 
the entailment label is explicitly known:  as such, this provides a ripe setting for \emph{interventional} studies
on the NLI models' representations, allowing for stronger causal conjectures and a 
deeper critical analysis 
of interventional probing methods.
In this work, we carry out new and existing representation-level interventions to investigate
the effect of these semantic features on NLI classification: we 
perform \emph{amnesic} probing (which removes features as directed by learned linear probes)
and introduce the \emph{mnestic} probing variation (which forgets all dimensions
\emph{except} the probe-selected ones).
Furthermore, we delve into the limitations of these methods and outline some pitfalls  have been obscuring the effectivity of interventional probing studies. 

\end{abstract}

\section{Introduction}

The \emph{probing} paradigm has emerged as a useful interpretability methodology which has been shown to have reasonable information-theoretic 
underpinnings \cite{pimentel_itprobing, voita-titov, zhu_probing}, 
indicating whether a given feature is captured in the intermediate
vector representations of neural models.
It has been noted many times that this does not generally imply that
the models are \emph{using} these learnt features, and they may represent 
vestigial information from earlier training steps \cite{ravichander_probing, amnesic}.

Only through interventional analyses can we start to make claims
about which modelled features are used for a given downstream task:
this is the aim of works such as \citet{amnesic, giulianelli_hood}
and \citet{geiger_causal}. 
We refer to the case where the interventions are guided by trained probes as \emph{interventional probing}.

\begin{figure}
    \centering
    \includegraphics[trim={1cm 12cm 2cm 0cm},clip,width=\columnwidth]{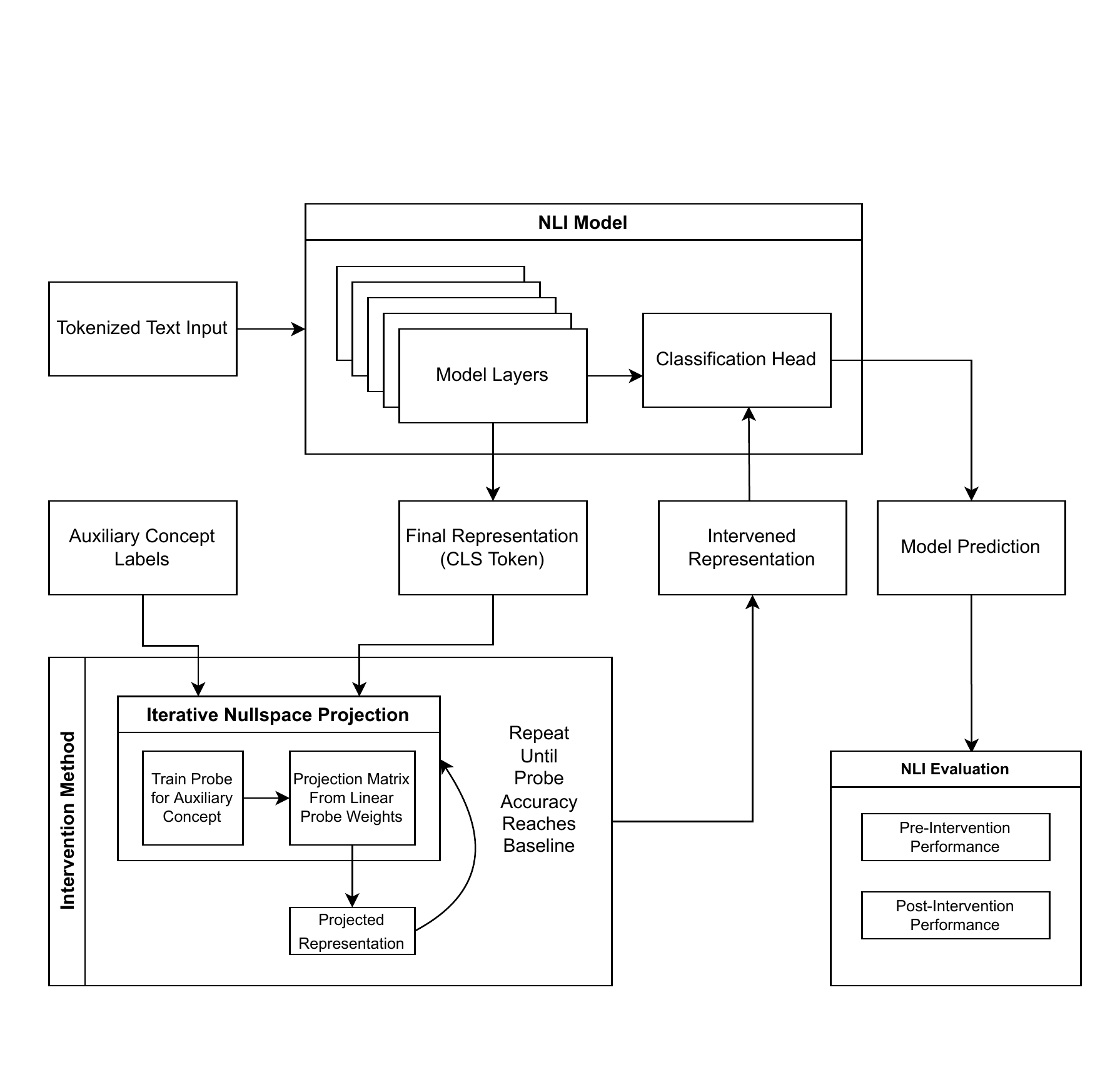}
    \caption{Workflow for interventional probing for NLP classification models: a basis for both the \emph{amnesic} and \emph{mnestic} intervention strategies.}
    \label{fig:probing_flow}
\end{figure}

It has been suggested in \citet{amnesic} (as the guidance for their \emph{amnesic probing} methodology) 
that if features are strongly detected by 
probes, one may use debiasing methods such as \emph{iterative nullspace projection (INLP)}
\cite{ravfogel_nullspace}
to intervene on the corresponding vector representations and effectively ``remove" 
the features before re-insertion into the given classifier. 
Investigating the effect of these intervention operations on the classifier performance could
allow for stronger causal claims about the role of the probe-detected features.

In this work, we delve deeper into the amnesic probing methodology 
with an NLI case study and identify two key limitations.
Firstly, there is an issue of dimensionality:
when the number of dimensions is high and the number of auxiliary feature classes
is low, it seems that amnesic probing is not sufficiently informative. In particular, we cannot rely on the same control baselines to reach the kind of conclusions
discussed in \cite{amnesic}, as nulling out small numbers of random directions    
consistently has no impact on the downstream performance.
Secondly, in the linguistic settings explored in \citet{amnesic},  we do not have 
expectations for exactly \emph{how} or even \emph{if} the explored features should 
be affecting the downstream task.
This makes it difficult to explore the effectivity of the methodology itself.

To this end, we propose the use of a controlled subset of NLI called
\emph{natural logic} \cite{maccartney-manning}. 
In this setting, the intermediate linguistic features of \emph{context 
montonicity} and \emph{lexical relations}
are already known to be highly extractable from certain NLI models'
hidden layers \cite{rozanova_decomposing}, allowing us a certain
amount of understanding and control of these features' representations in the latent space. 
Using the deterministic and well-understood nature of the 
problem space where we have concrete \emph{expectations} about the theoretical
interaction between the intermediate features and the downstream label, 
we may critically analyse the effectivity of interventional probing.

Through the application of probe-based interventions in this setting, 
we show that blindly applying the amnesic probing argument structure leads to 
unexpected and contradictory conclusions:
the two features which the final label is \emph{known} to depend on are shown to 
have no influence on the final classification (both jointly and independently).
This further calls into question the suitability of these methods for situations where
a small number of feature label classes and high dimensionality of representations is 
concerned. Even more perplexingly, when we treat the NLI gold label itself as an 
intermediate feature which can be nulled out with INLP, we yet again observe 
\emph{almost no change to the NLI performance}.  As such, the feature removal strategy appears ineffective here: we attribute this 
to the disproportionate size of probe-selected feature subspaces 
to the very high-dimensional representations. 

In response, we introduce and study a variation which we call 
\emph{mnestic} probing, which we show to be more informative in the high-dimensional, low-class-count setting: the core idea is to \emph{keep only} the directions
identified by the iteratively trained probes.
This allows us to analyse much lower dimension subspaces, while making better use of the outputs of the INLP strategy used in amnesic probing. 

We find that \emph{mnestic probing} leads to more informative observations which are a) in line with expected behaviour for
natural logic, and b) yield results which seem to better discriminate between model behaviours.

In summary, the contributions of the paper are as follows:
\begin{enumerate}
    \item We propose the setting of \emph{natural logic} to be ripe territory 
    for exploration of interventional probing strategies.
    \item We note two limitations of the amnesic probing methodology, demonstrating 
    both dimensionality limitations for the control baselines \ref{sec:controls} 
    and contradictory behaviour in the NLI setting \ref{sec:results} (namely that 
    that the expected effects of semantic features on the downstream NLI task are notably absent).
	\item Building upon previous interventional methodologies, we introduce an additional \emph{mnestic} intervention operation which uses the outputs of the INLP process in the opposite way. 
	\item We contrast the mnestic probing strategy with the amnesic probing results, and demonstrate it presents more informative results which are aligned with the constructed expectations in our high dimensional, low label class
	count setting.
\end{enumerate}


\section{Interventional Probing}\label{sec:interventions}

We may summarise the general setup of interventional probing as follows:
suppose we start with a classification model that may be decomposed as 
$f\circ g : \mathcal{X} \to \mathbb{R}^n$, where $g$ is an encoder module
which yields a representation which serves as an input to the classifier head $f$, 
and $n$ is the number of output classes of the final classifier.
We aim to intervene on the output of $g$ and observe the change in the performance of 
$f$ (usually in comparison with some kind of random control baseline intervention).

Linear probes (also known as \emph{diagnostic classifiers}) are able to identify subspaces in which a given intermediate feature set is 
found to be represented.
These may be used as a guide for vector-level interventions on the
representation space; we are specifically concerned with interventions which are vector \emph{projections}. Otherwise, The exact nature of this intervention is interchangeable. 
We consider two projection strategies in 
particular: 
the \emph{amnesic} intervention introduced in \citet{amnesic} 
(described further in section \ref{sec:amnesic}) and our 
\emph{mnestic} variation which uses the same INLP technique (section \ref{sec:mnestic}).


\subsection{What Should it Tell Us?}

The interventional probing steps are performed on exactly the representation
that would have been an input to the classifier head $f$. 
We may re-insert the intervened representations and re-calculate the 
classifier accuracy (note that the iterative projections in sections \ref{sec:amnesic} and 
\ref{sec:mnestic} maintain the original dimensionality of the vector set 
but reduce the \emph{rank}). 

We are looking to see if the downstream performance of the classifier $f$ drops.
If it does, the interventions have removed information that was necessary for 
successful classification.
However, as any projection would remove some information, these results must be
viewed in the context of a control intervention: if the INLP process ends up
removing $n$ directions, a sample of $n$ randomly chosen directions is 
selected from the original representation,
\citet{amnesic} argue that if the amnesic downstream performance drops significantly
more than the random removal control performance, we may conclude that the features 
were necessary for the final downstream classification. On the other hand, if the performance does not drop at all, the features were not useful for the classifier in 
the first place. 
In the ensuing sections and results, we demonstrate that this is not necessarily a valid 
conclusion.
    

\subsection{The Amnesic Intervention}\label{sec:amnesic}
We follow the procedure in \cite{amnesic} (in turn based on \emph{iterative nullspace projection} \cite{ravfogel_nullspace}):
given a set $X$ of encoded representations for the textual input 
(with dimensions \textttt{num\_examples} $\times$ \texttt{embedding\_dimension}), 
we iteratively train linear SVM classifiers according to a set of auxiliary feature labels. 
For each INLP step $i$, This yields a linear transformation $W_iX + B$, where the vectors of $W_i$ define directions onto which the probe projects the representations for auxiliary label classification (i.e., these are the
chosen directions most aligned with auxiliary class separation). 
For each step $i$, an orthogonal basis denoted $R_i$ is found for this rowspace.
The projection to the intersection of the nullspaces is given by a matrix 
$$PX = (I - (R_0 + ... + R_n)) X.$$
The matrix product $PX$ is a matrix in the original dimensions of $X$, but with reduced rank
by the number of iteration steps (as each projection "flattens out" the representation in these directions).

Projection to the intersection of nullspaces is thus the removal of any information 
pertaining to the auxiliary feature labels (or at least, the information which allows high 
performance for a linear probe). 
The training terminates these auxiliary task classifiers start 
consistently performing at the majority class baseline, indicating that there 
is no further linearly information to be extracted from the remaining 
representation.
As such, the resulting representation is treated as an altered representation where this feature is \emph{removed} or forgotten.
    
\subsection{A Variation: The Mnestic Intervention}\label{sec:mnestic}


\citet{amnesic} perform a series of experiments on various linguistic features which
had previously been shown to be well-captured in language model representations 
and use the amnesic probing methodology to distinguish between
features that are \emph{used} by the model and those that are not by 
comparing post-intervention downstream task performance to a baseline of randomly
removed directions.


Rather than projecting the embedded representations to the intersection of nullspaces of the
trained probes (removing the target property), we project them to the \emph{union of the rowspaces}
with the transformation:
\begin{align*}
    (I-P)X &= (I - (I - (R_0 + \ldots + R_n))) X \\
            &= (R_0 + \ldots + R_n) X 
\end{align*}

This has the opposite effect: 
we use projection to null out \emph{everything except} the directions
identified by the probes as indicative of the target feature.
As such, we "remember" only that feature rather than forgetting it.


\section{Experimental Setup}
In this study, we use interventional methods
\footnote{We reuse much of the code included with \cite{amnesic}, but we include our
data and reproducible experimental code at \url{https://github.com/juliarozanova/mnestic_probing}.} to study the internal behaviour
of NLI models.
We compare amnesic and mnestic variations of the INLP strategy, evaluating 
intermediate feature probing performance and downstream NLI performance after 
every step of the intervention process.


For each auxiliary feature label and and model, we perform the 
\emph{interventional probing} as outlined in figure \ref{fig:probing_flow}.

\subsection{Dataset}
Our setting for this study is a fragment of NLI called \emph{Natural Logic} 
\cite{maccartney-manning}. In particular, we focus on single-step natural logic
inferences in which entailment examples are generated by replacing a noun phrase
in a sentence with a hyponym, hypernym or unrelated noun phrase.
The context of the substituted term is either \emph{upward} or \emph{downward} 
monotone, as determined by the composition of negation markers, generalized
quantifiers or determiners present in the context. 
The entailment label of the example is a consequence of this feature and the lexical 
relation between the substituted terms.

\begin{figure}[h!]
    \centering
    \resizebox{\columnwidth}{!}{
    \begin{tikzpicture}
        \node[state] (1) at (0,0) {Context Monotonicity};
        \node (2) [right = of 1] {};
        \node[state] (3) [right = of 2] {Lexical Relation};
        \node[state] (4) [below = of 2] {Entailment Label};
    
        \path (3) edge (4);
        \path (1) edge (4);
    \end{tikzpicture}%
    }
    \caption{}
    \label{fig:data_dep}
\end{figure}

We use the NLI\_XY dataset from \cite{rozanova_decomposing, rozanova_supporting}.
By construction, the NLI\_XY dataset consists of NLI examples
which rely on exactly these two abstract features: context monotonicity and the lexical relation of the substituted terms.

We perform two flavours of probe-based interventions (described fully in section \ref{sec:interventions}) with four feature label sets (described next).

\paragraph{Auxiliary Feature Labels}
We begin with the two relevant intermediate features 
(respectively, context monotonicity and 
lexical relation) which are already known to correlate with stronger
performance on the downstream NLI\_XY task \cite{rozanova_decomposing}.
We will refer to this as \emph{single-feature} interventional probing, as 
the probing and intervention steps are only applied to one feature set at 
a time.
Next, we combine the two features in a cross product, creating a new feature 
label set with all possible combinations of these intermediate features 
(in the dataset, they are completely independent variables by construction 
\cite{rozanova_supporting}). We refer to this as the \emph{composite feature label}.

Lastly, we also consider the \emph{entailment label} itself (the downstream
task label) as an input to the interventional probing process.
The latter is particularly useful as a diagnostic sanity check, 
and aids the critical nature of our findings.

\subsection{NLI Models and Encoding}

We compare a selection of \textttt{BERT} \cite{bert} and \textttt{RoBERTa} \cite{roberta} models trained for 
NLI classification. 
Firstly, we include a pair of models trained respectively on the MNLI \cite{mnli}
and SNLI \cite{snli} benchmark datasets.
In \cite{rozanova_decomposing} and \cite{rozanova_supporting}, it is shown that 
when \textttt{roberta-large-mnli} (a model which performs well on benchmarks
but poorly on the targeted NLI\_XY challenge set) receives additional training
on the adversarial HELP dataset \cite{yanaka_help} it improves in NLI\_XY 
performance
and \emph{begins to show high probing performance for the relevant intermediate 
features}, context monotonicity and lexical relations: this is the necessary
precondition for doing interventional probing.
We include two of their models with this property: 
\textttt{roberta-large-mnli-help} and 
\textttt{roberta-large-mnli-double-finetuning}, with the other models included 
for a contextual comparison.

We perform probing and intervention on the final representation that precedes the
NLI classification head: in the case of \textttt{BERT} and \textttt{RoBERTa}, 
this is the \textttt{[CLS]} token of the final layer.  


The initial input is a tokenized NLI example from the NLI\_XY dataset.
The findings in \cite{rozanova_decomposing} show that the intermediate feature
labels (context monotonicity and lexical relations) are detectable in the 
concatenated tokens of the substituted noun phrases:
however, for interventional purposes, we perform the probing and intervention
steps on the \textttt{[CLS]} token which serves as an input to the NLI classifier
head: we have found that the same features are detectable to a 
comparable standard, and this is the only position at which we are able to make
a sensible intervention that would allow conclusions about the final classifier
head only.

\subsection{Evaluation}
The significant metrics for these interventional probing paradims are the
\emph{probing accuracy} before and after the iterative nullspace projection steps 
(a decline to random performance indicates the feature is being ``removed" from 
the representation in the sense that it is no longer detectable by linear 
probes) and the \emph{downstream classification accuracy} on the NLI task the 
model's were trained for (in our case, we report the accuracy on the NLI-XY  
task).

\begin{table*}[h!]
    \centering
    \resizebox{0.9\textwidth}{!}{
    \begin{tabular}{@{}llrrrr@{}}
    \toprule
                                         &                      & \multicolumn{2}{l}{Probing Performance}                                              & \multicolumn{2}{l}{NLI-XY Performance}                                               \\
    \textbf{Model}                       & \textbf{Feature}     & \multicolumn{1}{l}{\textbf{Start}} & \multicolumn{1}{l}{\textbf{Intervention $\Delta$}} & \multicolumn{1}{l}{\textbf{Start}} & \multicolumn{1}{l}{\textbf{Intervention $\Delta$}} \\ \midrule
    roberta-large-mnli-help              & insertion relation   & 80.58                              & -40.35                                          & 79.79                              & 0.06                                            \\
    
                                         & context monotonicity & 87.65                              & -46.22                                          & 79.79                              & -0.09                                           \\
                                         & composite            & 64.48                              & -43.95                                          & 79.79                              & 0.32                                            \\
                                         & entailment label     & 78.05                              & -37.49                                          & 79.79                              & -1.57                                           \\ \midrule
    roberta-large-mnli-double-finetuning & insertion relation   & 62.7                               & -36.49                                          & 80.04                              & 0.11                                            \\
                                         & context monotonicity & 89.79                              & -43.28                                          & 80.19                              & 0                                               \\
                                         & composite            & 57.64                              & -49.56                                          & 80.08                              & -1.67                                           \\
                                         & entailment label     & 82.8                               & -24.94                                          & 80.19                              & -16.53                                          \\ \midrule
    roberta-large-mnli                   & insertion relation   & 80.39                              & -45.59                                          & 57.22                              & 8.99                                            \\
                                         & context monotonicity & 75.44                              & -27.49                                          & 57.37                              & -0.43                                           \\
                                         & composite            & 72.35                              & -53.51                                          & 57.24                              & -2.27                                           \\
                                         & entailment label     & 73.6                               & -15.31                                          & 57.37                              & 0.1                                             \\ \midrule
    bert-base-uncased-snli-help          & insertion relation   & 59.53                              & -19.1                                           & 45.95                              & 0.28                                            \\
                                         & context monotonicity & 82.72                              & -33.94                                          & 45.52                              & -2.35                                           \\
                                         & composite            & 37.19                              & -17.08                                          & 45.76                              & 13.68                                           \\
                                         & entailment label     & 47.05                              & 0.38                                            & 45.91                              & 0                                               \\ \midrule
    bert-base-uncased-snli               & insertion relation   & 60.26                              & -35.14                                          & 48.99                              & 1.05                                            \\
                                         & context monotonicity & 81.09                              & -30.77                                          & 49.42                              & -6.25                                           \\
                                         & composite            & 35.37                              & -17.83                                          & 50.73                              & 7.45                                            \\
                                         & entailment label     & 42.44                              & -0.24                                           & 49.42                              & 0                                               \\ \bottomrule
    \end{tabular}%
    }
    \caption{Amnesic probing performance deltas across models and target feature labels: first listed is the performance on the probing task with respect to the indicated feature, and then the 
    accuracy on the downstream NLI-XY task. We note the results pre-intervention and the ensuing change in accuracy.}\label{table:results}
  \end{table*}

For amnesic probing, we report the performance deltas for both 
the probing and downstream tasks.
However, for mnestic probing, a slightly more nuanced and qualitative view is helpful: it can be assumed that eventually
mnestic probing will reach comparable performance to the untouched
vector representations, but we are interested in the comparative rates at which this happens.
As the interventions are iterative, we may feed the intervened representations into the classifier head at \emph{each step}
of the intervention process - we use this to provide a step-wise presentation of 
results in linear plots in figure \ref{fig:single_mnestic}.

While the tabulated deltas in table \ref{table:results} results
are sufficient to present our observations on amnesic probing, for comparison we also include the stepwise graphical presentations in the appendix. 

\section{Results and Discussion}

\subsection{Single Feature Amnesic Probing}

The results for the standard amnesic probing procedure are in table \ref{table:results}.  
In particular, the single feature results are in the rows 
with features labelled \emph{insertion relation} and \emph{context monotonicity}.
The amnesic operation is successful - the respective probing accuracies approach 
and reach the majority class baseline. 

\begin{figure}[h!]
    \centering
    \includegraphics[width=0.48\textwidth]{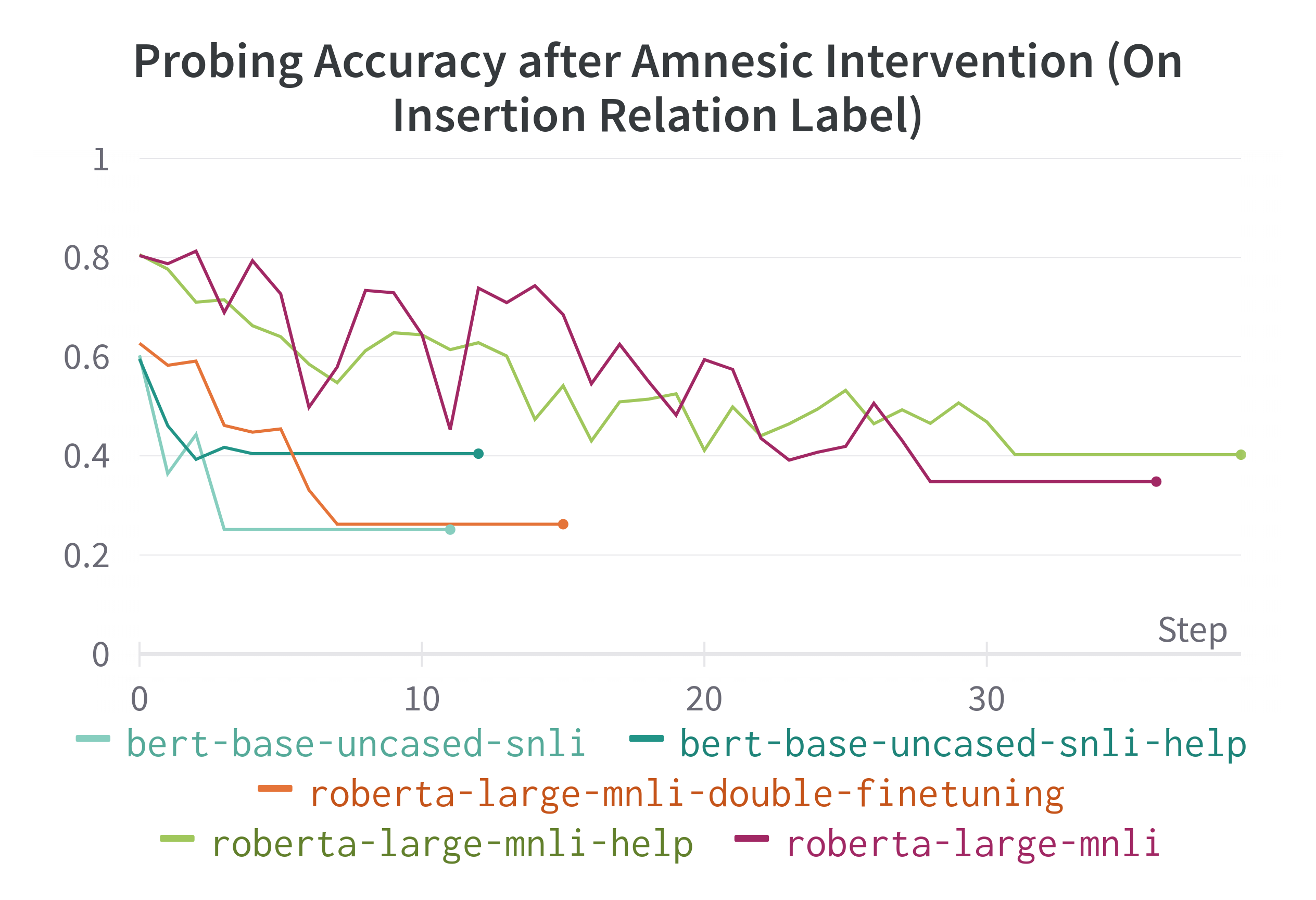}
  \caption{Step-wise probing performance throughout the amnesic probing process: a decrease towards the random baseline accuracy (roughly 0.3 for this 3-class task) indicates the feature is less and less extractable from the remaining representations as the iterative process continues.}
    \label{fig:in_text_amnesic_ins}
\end{figure}

We also include the step-wise plots of both probing performance and downstream NLI 
task performance: we single out the case of the insertion relation label in figures 
\ref{fig:in_text_amnesic_ins} and \ref{fig:in_text_nli_xy_amnesic_ins}, but include
the full suite of expanded plots for each feature in the appendix.
The length of the iterative amnesic probing process is indicative of the number of dimensions 
removed to reach this baseline: it can also be considered a proxy for the strength
of the feature presence in the representations, or rather, the dimension of the 
semantic subspace corresponding to the target features.

\begin{figure}[h!]
    \centering
    \includegraphics[width=0.48\textwidth]{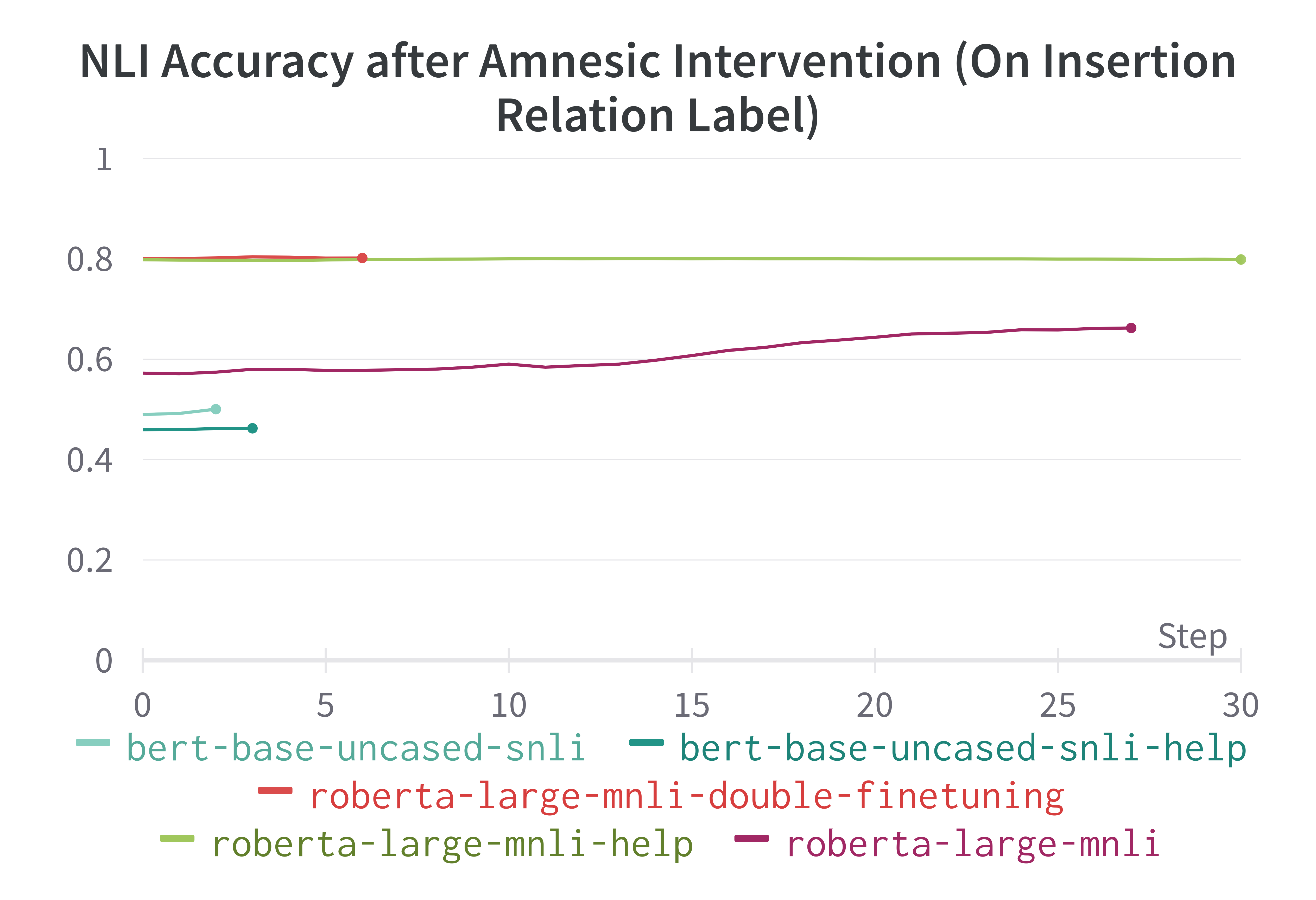}
  \caption{Downstream performance on NLI\_XY after amnesic intervention (removing lexical relation information). For such an important feature to the end-task, we would expect to see a drop: but we don't!}
    \label{fig:in_text_nli_xy_amnesic_ins}
\end{figure}

The second phase of this process, i.e. the resubstitution of the modified
representations as inputs to the NLI classifier head, can be seen in
the right hand portion of table \ref{table:results}, labelled \emph{NLI-XY Performance}.
The result is unexpected: for each of these features, 
\emph{the downstream task performance appears to be unaffected after their removal.}
This is surprising when the dataset is explicitly controlled to rely only on these two features. 



\subsection{Multi Feature Amnesic Probing}\label{sec:results}
The results for the amnesic probing procedure utilizing \emph{both} auxiliary
feature label sets and the entailment gold label are in the rows of table \ref{table:results} with labels \emph{composite} and \emph{entailment label} respectively.
We observe that once again, the downstream task performance is mostly unaffected.
Unlike the unexpected result in the previous section, it's difficult to argue
away the fact that this is somewhat contradictory: while single feature removal may be subject to some confounding bias, the removal of both
features exhausts the variables on which this classification depends.
This is highly unexpected, and suggests a point of failure for the amnesic probing process.
Naturally, we cannot be without doubt that despite all our best 
efforts to work with a controlled dataset that relies only on these two know (but still complex) features, a model may yet 
find unrelated heuristics to exploit that may correlate so 
strongly with the downstream task label that it may perform well
without representing and using these intermediate features.
However, we imagine this to be a rather low probability scenario to be that the model simultaneously learns such heuristics but simultaneously learn representations that create strong clusters for the known intermediate features \emph{without using them at all}. 
The models which we have observed to perform more less well on NLI-XY (such as roberta-large-mnli) are indeed estimated to be using sub-par heuristics, but this also comes with poor probing results for the intermediate features - naturally, this in itself does not imply anything conclusive, but certainly adds to our convictions.

On a seprate note, it is noted in \citet{amnesic} that there is no control for the number of 
dimensions removed, while there is a clear correlation between downstream task performance 
and the number of label classes (and thus removed probe directions) are in play.
Our feature sets have only 2 and 3 classes respectively. In the most analagous result in \cite{amnesic} where the 
auxiliary features had very few classes and no change on the downstream performance was observed, it
was concluded that the features must have no effect on the outcome.
It is very likely that \emph{too little information} is being removed in this process to observe any
impact on the downstream task performance. This could potentially be pointing to high redundancy in the representations which the amnesic intervention may struggle to remove appropriately.

\subsection{Mnestic Probing}

\begin{figure*}[h!]
\centering
\begin{subfigure}{.5\textwidth}
  \centering
  \includegraphics[width=\textwidth]{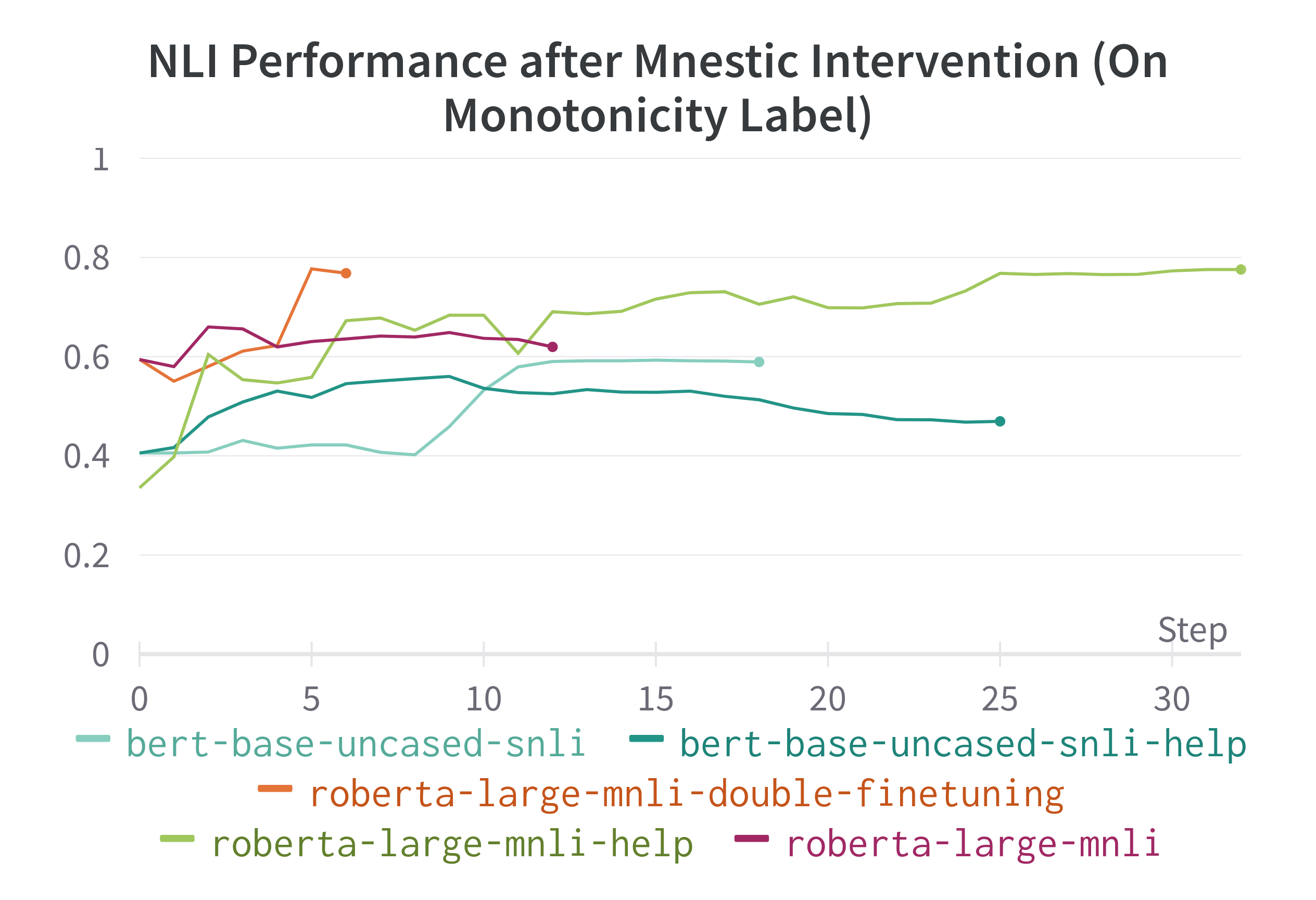}
  \caption{Context Monotonicity Label}
\end{subfigure}%
\begin{subfigure}{.5\textwidth}
  \centering
  \includegraphics[width=\textwidth]{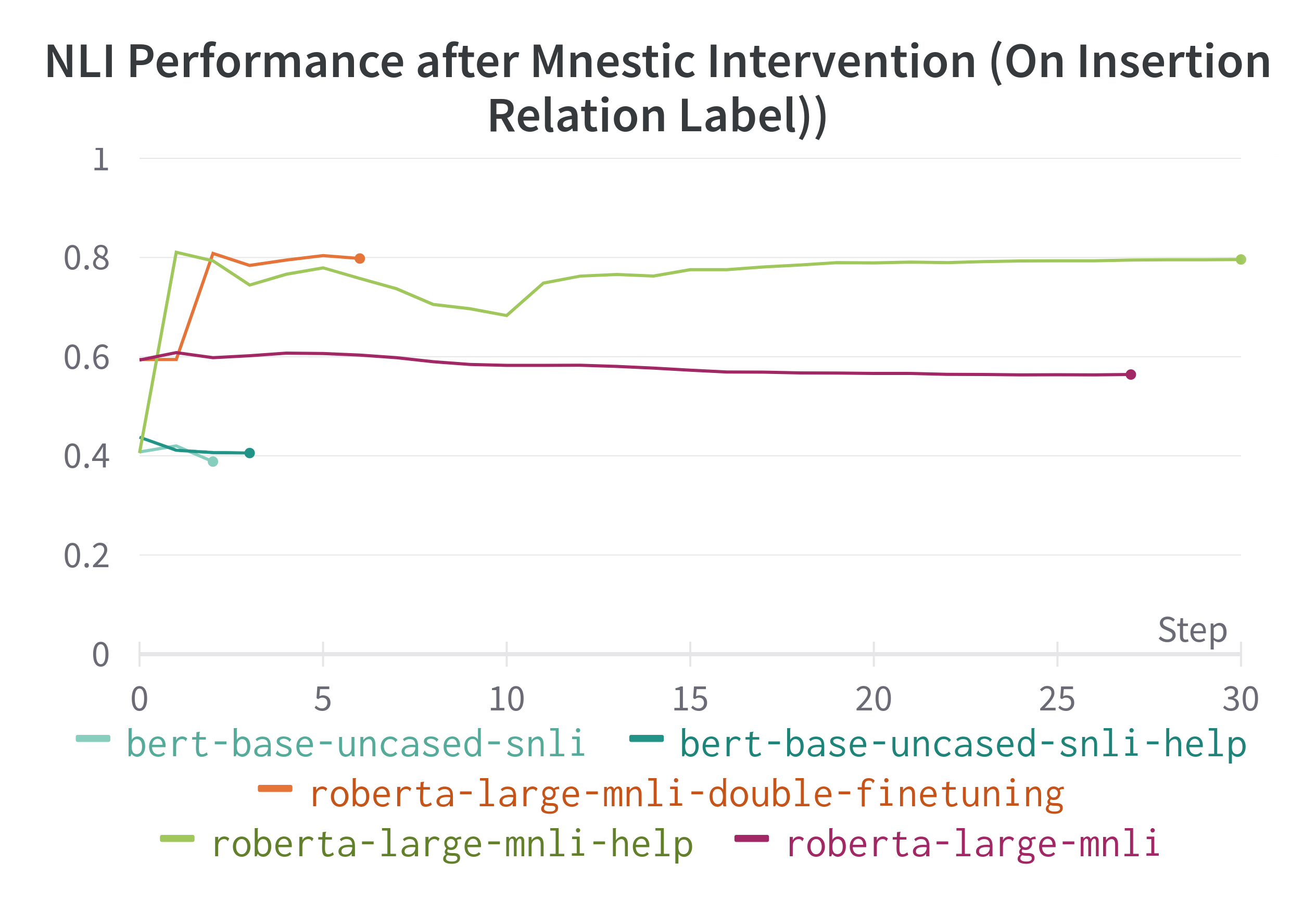}
  \caption{Lexical Relation Label}
\end{subfigure}
\begin{subfigure}{.5\textwidth}
  \centering
  \includegraphics[width=\textwidth]{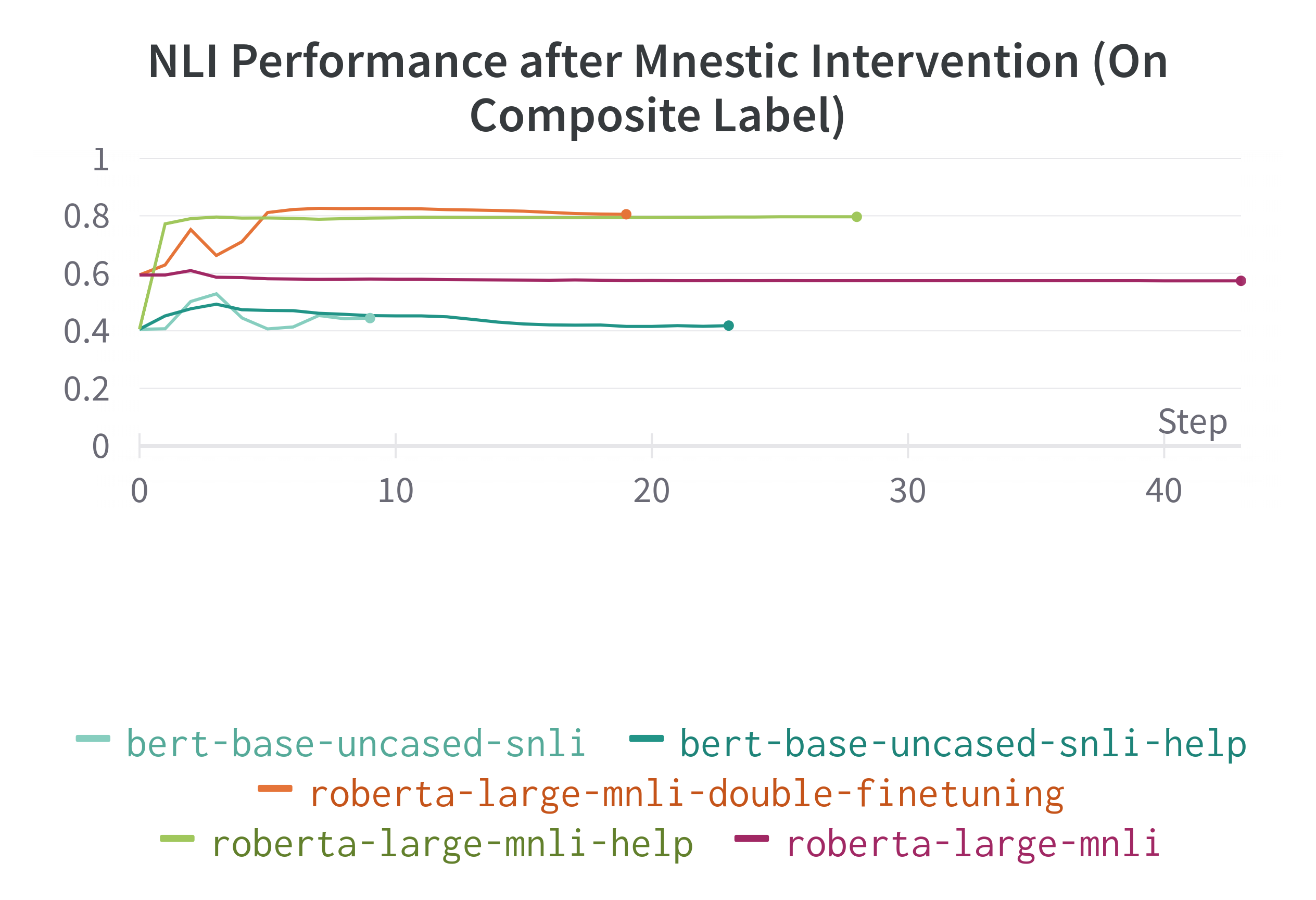}
  \caption{Composite Label}
\end{subfigure}%
\begin{subfigure}{.5\textwidth}
  \centering
  \includegraphics[width=\textwidth]{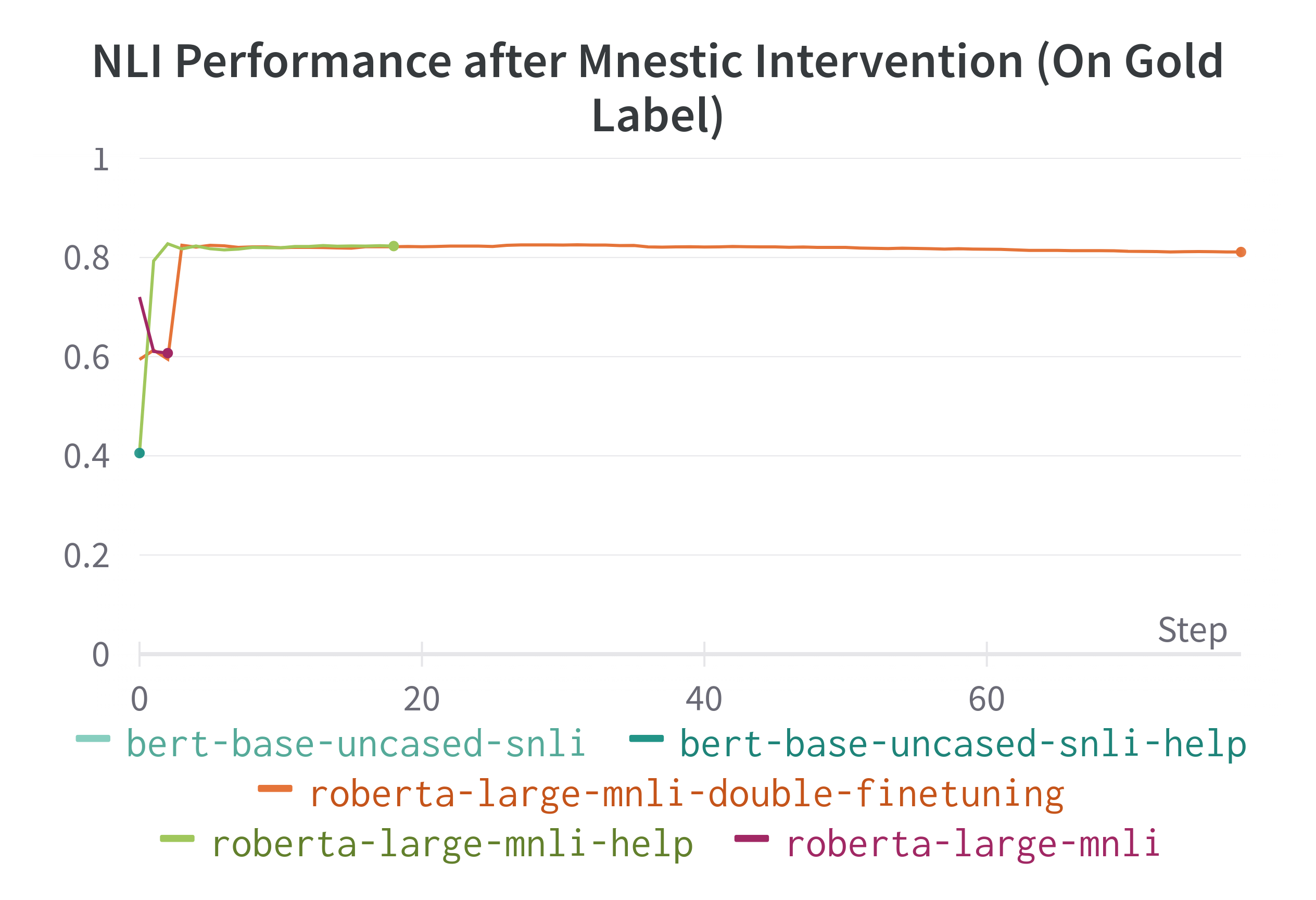}
  \caption{Gold Label}
\end{subfigure}
\caption{Downstream NLI Task Performance After Mnestic Interventions}
\label{fig:single_mnestic}
\end{figure*}

Given the possible dimensionality problem, the alternative method of \emph{mnestic} probing seems 
promising: after the mnestic intervention, many dimensions are removed and few remain, so it appears to be a ripe setting
for observing and comparing effects on downstream NLI accuracy at a finer granularity.
The results for NLI-XY task accuracy after the \emph{mnestic} probing procedure are presented as step-wise plots in figure \ref{fig:single_mnestic}.
There is a clear increase in NLI performance with subsequent addition
of probe-chosen directions to the representations, especially viewed in the context of
section \ref{sec:controls}, where we compare the performance to random 
choices of included directions. In the latter, performance varies randomly rather than presenting a structured increase as seen here.

We observe that the \emph{composite} label and the gold \emph{entailment} label
are reflected in line with expectations in the mnestic probing experiments: the inclusion of 
the probe-selected dimensions with respect to these labels introduces a sharp and
immediate increase in the NLI classifier performance. This is significantly 
steeper than the baseline increase observed in random addition of representation
directions.
Similarly, the increase is nearly as sharp for the lexical relation label.
However, although an increase is observed during the iterative mnestic probing 
intervention for context montonicity, this increase is not at a dramatically higher
rate than adding subsequently more directions from the original representation.
For monotonicity specifically, this is not enough to conclude that the feature 
(or at least, the corresponding probe-selected dimensions) are critical to the 
final classifier.

Nevertheless, we have been able to make clearer observations than were possible in 
the amnesic probing setting.

\subsection{Control Comparison} \label{sec:controls}

\begin{figure}[h!]
    \centering
    \resizebox{\columnwidth}{!}{
    \includegraphics{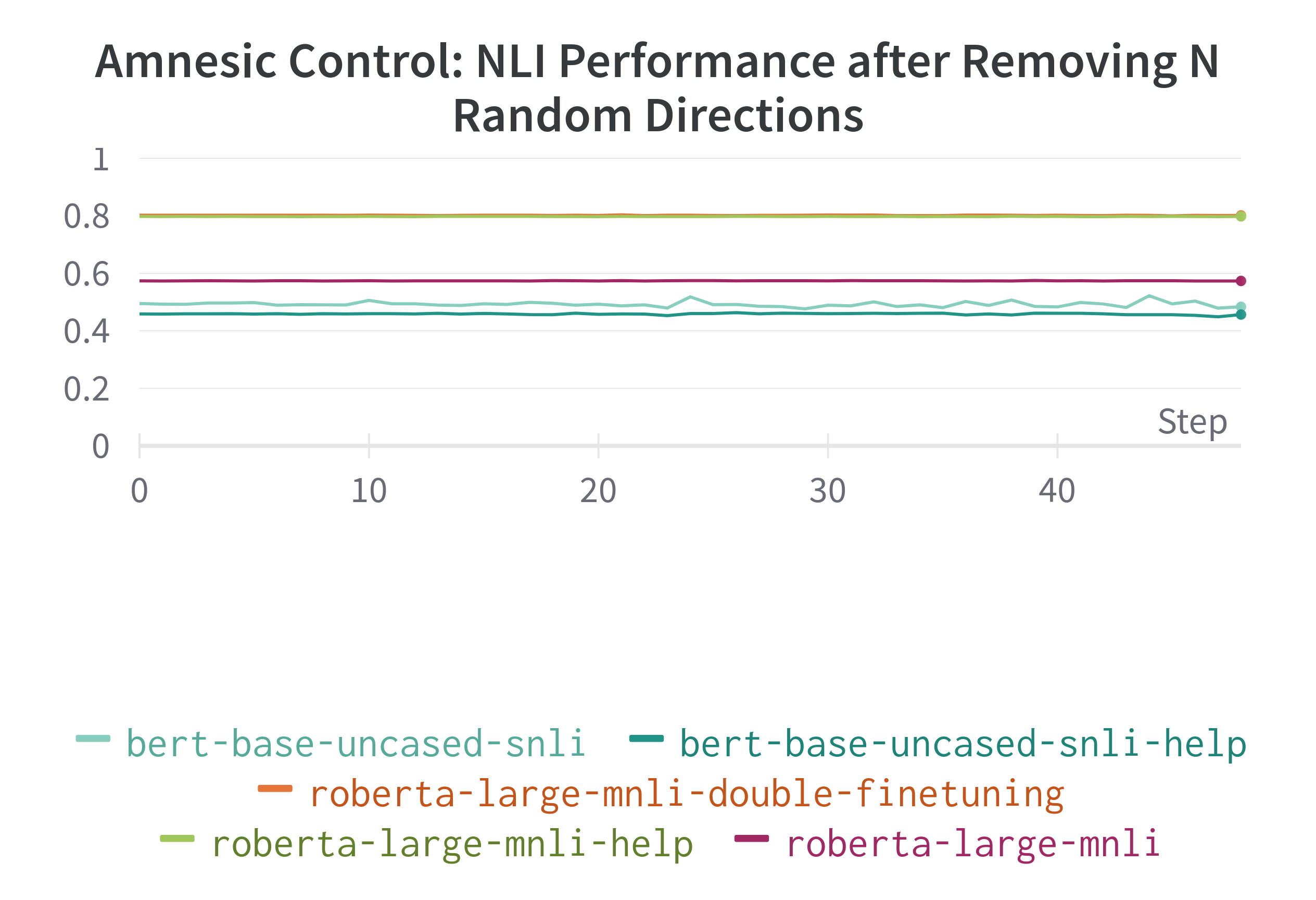}%
    }
    \caption{Amnesic control experiment: Downstream NLI accuracy upon the \emph{removal} of $n$ random directions of the original representation.}
    \label{fig:control_amnesic}
\end{figure}

\begin{figure}[h!]
    \centering
    \resizebox{\columnwidth}{!}{
    \includegraphics{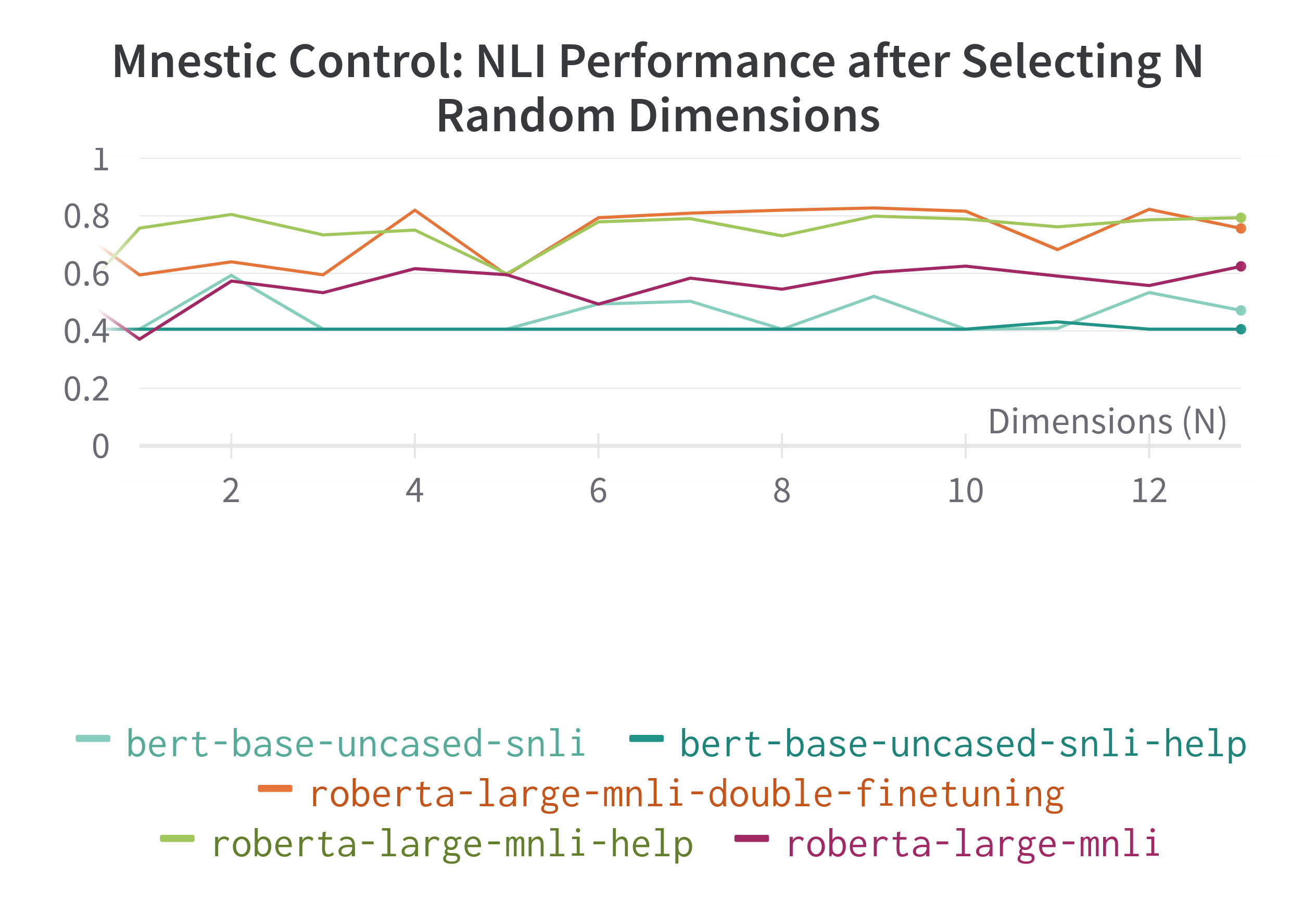}%
    }
    \caption{Mnestic control experiment: downstream NLI accuracy upon the \emph{selection} of $n$ random directions of the original representation.}
    \label{fig:control_mnestic}
\end{figure}
We contextualise all the preceding results with a set of control experiments
both for amnesic (figure \ref{fig:control_amnesic}) and mnestic (figure \ref{fig:control_mnestic})
probing.
Note in particular that even with very few random dimensions kept,
downstream performance starts approaching comparable levels to the full
representations.
As such, a single random baseline as in \citet{amnesic} can be misleading: there is 
enough variability in the random direction results so as to allow for a false claim of
feature irrelevance by simply getting lucky; as few as 3 dimensions can perform at 
the original model's performance level or arbitrarily lower.

Lastly, we compare to the mnestic probing results in figure \ref{fig:single_mnestic}:
with the probe-selected mnestic dimension choices, the increase in downstream
performance does seem to happen faster and in a more consistent fashion,  
while the selection of $n$ randomly chosen directions introduces very haphazard performance spikes.
This suggests the probe-selected dimensions are consistently adding to the 
model's access to the relevant information, amd this may be stronger evidence for the 
usefulness of the examined features for the final classification.

\section{Related Work}

\begin{table*}[ht!]
\resizebox{\textwidth}{!}{
\begin{tabular}{@{}l|llllll@{}}
\toprule
                                                                                                                    & Intervention                                                                                        & Tested Effect                                                                                            & Feature Characterisation                                       & \begin{tabular}[c]{@{}l@{}}Requires Intermediate \\ Labels\end{tabular} & \begin{tabular}[c]{@{}l@{}}Intervention Linked to\\ Concept Interpretation\end{tabular} & Domain                \\ \midrule
Amnesic Probing / INLP \cite{amnesic}                                                                                              & Debiasing / Feature Removal                                                                         & Downstream Classifier Accuracy                                                                           & Linear Classifier                                              & Yes                                                                     & No                                                                                      & Language Modelling    \\
\\
\begin{tabular}[c]{@{}l@{}}CausaLM: Causal Model Explanation \\ Through Counterfactual Language Models\end{tabular} \cite{causalm} & \begin{tabular}[c]{@{}l@{}}Re-Training Model Copy \\ For Counterfactual Representation\end{tabular} & \begin{tabular}[c]{@{}l@{}}Text representation-based individual\\ treatment effect (TReITE)\end{tabular} & \begin{tabular}[c]{@{}l@{}}Retrained Base\\ Model\end{tabular} & Yes                                                                     & Yes                                                                                     & Sentiment Analysis    \\

\\
\begin{tabular}[c]{@{}l@{}}Explaining Classifiers\\ with Causal Concept Effect\end{tabular} \cite{cace}                                                                                              & Generative Modeling                                                                                 & Average Causal Effect Measure                                                                            & VAE                                                            & Yes                                                                     & Yes                                                                                     & Vision Classification \\
\\
Concept Activation Vectors (TCAV) \cite{tcav}                                                                                   & Value Shift in Vector Direction                                                                     & Custom Gradient Sensitivity Measure                                                                      & Linear Classifier                                              & Yes                                                                     & Yes                                                                                     & Vision Classification \\
\\
\begin{tabular}[c]{@{}l@{}}Latent Space Explanation \cite{latentspace} \\ by Intervention\end{tabular}                                 & \begin{tabular}[c]{@{}l@{}}VAE Input Discretization \\ and Reconstruction\end{tabular}              & Reconstruction Quality                                                                                   & VAE                                                            & No                                                                      & \begin{tabular}[c]{@{}l@{}}Qualitative Judgement \\ (Vision Only)\end{tabular}          & Vision Classification \\
\\
\begin{tabular}[c]{@{}l@{}}Meaningfully Debugging Model Mistakes \\ Using Conceptual Counterfactuals\end{tabular} \cite{meaningful}  & \begin{tabular}[c]{@{}l@{}}Weighted Combination of\\ Concept Vectors\end{tabular}                   & \begin{tabular}[c]{@{}l@{}}Difference Between Concept\\ Addition and Removal Effect\end{tabular}         & Linear Classifier                                              & Yes                                                                     & Yes                                                                                     & Vision Classification \\ \bottomrule
\end{tabular}%
}
\caption{Related Work on Latent Concept Interventions}\label{table:lit}
\end{table*}

The use of probing as an interpretability strategy dates back as far as 
works such as \citet{alainbengio} and \cite{conneau_cram}, but a core set of work 
on the detailed development of the methodology includes \citet{hewitt-liang, belinkov_glass, voita-titov, pimentel_itprobing}. For a full survey, see \citet{belinkov_probingsurvey}.

The application of probing strategies to natural logic components has been 
explored in \citet{rozanova_decomposing} and \citet{geiger}.
In \citet{rozanova_decomposing}, probing experiments have proven effective in detecting the presence or absence of features such as \emph{context monotonicity} and \emph{phrase-pair relations} in the internal representations of NLI models.

Regarding interventions as interpretability tools for machine learning classifiers, there are two broad categories: those that modify the raw input (such as image or text) in a controlled 
way, and those that modify the hidden/latent vector representations of the data 
at various stages of the models' input processing. 
While input-level interventions are more common as they are usually easier to 
control and are strongly interpretable,
they don't allow us to explore and conjecture about exact high-level 
representational mechanisms in the latent space.
We tabulate a few  relevant interventional interpretability methods in table 
\ref{table:lit}. Note in particular the variation in the \emph{generation} 
step for the intervened input; some use generative modelling for counterfactual examples, while we 
use cheaper linear probes.

The only other work in which interventional methods have been applied to natural logic
is \citet{geiger_causal}: a similar problem setting is considered,
but at a finer granularity. Our work focuses more on the summarised abstract notion
of context monotonicity as a single feature, rather than the intermediate tree nodes
that determine its final monotonicity profile.
The interventions used in this work are vector \emph{interchange} interventions; 
partial representations from transformed inputs are used, as opposed to direct
manipulations of the encoded vectors.


\section{Conclusion and Future Work}
Our expiremental setting has shown significant limitations of amnesic probing 
in high-dimensional settings where there are few label classes (and consequently
fewer dimension modified), even if these classes are strongly detectable.
Our results point out that it is misguided to concluded that a given feature
is not used when post-amnesic-intervention downstream performance fails to drop,
especially in our example amnesic probing studies of a) the gold donwstream 
feature label and b) the composite of two labels that jointly determine the 
entailment label.
This may be due to a dimension/rank confounder variable 
and high redundancy of information in the representations.
It remains to be checked whether high performance in the random control directions
corresponds to strong alignment with these probe-selected directions:
we propose an analysis of the \emph{dot products} with the fixed set of probe-selected
dimensions, which indicates a shared directionality measure ($0$ for orthogonal vectors and $1$ for
codirectional ones).

In summary: we have introduced a modification of the amnesic probing paradigm which
we call \emph{mnestic} probing which uses the same INLP process but considers
the opposite intervention: using the union of projection rowspaces to keep
\emph{only} the directions the probes have identified to be modelling the 
target information.
This strategy presents results that are more aligned with theoretical
expectations (in the NLI case), possibly because we are now able
to make comparisons in a lower rank setting.

\section{Limitations}
A key limitation of the mnestic probing strategy is that as one 
reconstructs the original representation one dimension at a time, information content
is naturally due to increase: as such, no mnestic probing result can be viewed in isolation, 
but should be used as a comparative study. Preferably, various randomized selections of 
linear subspaces with the same number of dimensions should be included as baselines input representations.
Furthermore, we mention two some additional caveats:
 firstly, the probing strategies used here to identify the informative semantic subspaces in question are always linear; relevant information may be present non-linearly. However,
    as with amnesic probing, we discount any non-linearly encoded information as the final
    model classifcation layer is linear and thus cannot exploit this information.
Lastly, probing for subspaces which are informative of target auxiliary features may always
    include correlated features in the resulting subspaces; this must always be taken into account when drawing conclusions from mnestic/amnesic probing.




    

\bibliography{anthology,custom}

\begin{thebibliography}{27}
\expandafter\ifx\csname natexlab\endcsname\relax\def\natexlab#1{#1}\fi

\bibitem[{Abid et~al.(2022)Abid, Yuksekgonul, and Zou}]{meaningful}
Abubakar Abid, Mert Yuksekgonul, and James Zou. 2022.
\newblock \href {https://proceedings.mlr.press/v162/abid22a.html} {Meaningfully
  debugging model mistakes using conceptual counterfactual explanations}.
\newblock In \emph{Proceedings of the 39th International Conference on Machine
  Learning}, volume 162 of \emph{Proceedings of Machine Learning Research}.
  PMLR.

\bibitem[{Alain and Bengio(2018)}]{alainbengio}
Guillaume Alain and Yoshua Bengio. 2018.
\newblock \href {http://arxiv.org/abs/1610.01644} {Understanding intermediate
  layers using linear classifier probes}.

\bibitem[{Belinkov(2022)}]{belinkov_probingsurvey}
Yonatan Belinkov. 2022.
\newblock \href {https://doi.org/10.1162/coli_a_00422} {Probing classifiers:
  Promises, shortcomings, and advances}.
\newblock \emph{Computational Linguistics}, 48(1):207--219.

\bibitem[{Belinkov and Glass(2019)}]{belinkov_glass}
Yonatan Belinkov and James Glass. 2019.
\newblock \href {https://doi.org/10.1162/tacl_a_00254} {Analysis methods in
  neural language processing: A survey}.
\newblock \emph{Transactions of the Association for Computational Linguistics},
  7:49--72.

\bibitem[{Bowman et~al.(2015)Bowman, Angeli, Potts, and Manning}]{snli}
Samuel~R Bowman, Gabor Angeli, Christopher Potts, and Christopher~D Manning.
  2015.
\newblock A large annotated corpus for learning natural language inference.
\newblock In \emph{EMNLP}.

\bibitem[{Conneau et~al.(2018)Conneau, Kruszewski, Lample, Barrault, and
  Baroni}]{conneau_cram}
Alexis Conneau, German Kruszewski, Guillaume Lample, Lo{\"\i}c Barrault, and
  Marco Baroni. 2018.
\newblock \href {https://doi.org/10.18653/v1/P18-1198} {What you can cram into
  a single {\$}{\&}!{\#}* vector: Probing sentence embeddings for linguistic
  properties}.
\newblock In \emph{Proceedings of the 56th Annual Meeting of the Association
  for Computational Linguistics (Volume 1: Long Papers)}, pages 2126--2136,
  Melbourne, Australia. Association for Computational Linguistics.

\bibitem[{Devlin et~al.(2019)Devlin, Chang, Lee, and Toutanova}]{bert}
Jacob Devlin, Ming-Wei Chang, Kenton Lee, and Kristina Toutanova. 2019.
\newblock \href {https://doi.org/10.18653/v1/N19-1423} {{BERT}: Pre-training of
  deep bidirectional transformers for language understanding}.
\newblock In \emph{Proceedings of the 2019 Conference of the North {A}merican
  Chapter of the Association for Computational Linguistics: Human Language
  Technologies, Volume 1 (Long and Short Papers)}, pages 4171--4186,
  Minneapolis, Minnesota. Association for Computational Linguistics.

\bibitem[{Elazar et~al.(2020)Elazar, Ravfogel, Jacovi, and Goldberg}]{amnesic}
Yanai Elazar, Shauli Ravfogel, Alon Jacovi, and Yoav Goldberg. 2020.
\newblock \href {http://arxiv.org/abs/arXiv:2006.00995} {Amnesic probing:
  Behavioral explanation with amnesic counterfactuals}.

\bibitem[{Feder et~al.(2021)Feder, Oved, Shalit, and Reichart}]{causalm}
Amir Feder, Nadav Oved, Uri Shalit, and Roi Reichart. 2021.
\newblock \href {https://doi.org/10.1162/coli_a_00404} {{C}ausa{LM}: Causal
  model explanation through counterfactual language models}.
\newblock \emph{Computational Linguistics}, 47(2):333--386.

\bibitem[{Gat et~al.(2021)Gat, Lorberbom, Schwartz, and Hazan}]{latentspace}
Itai Gat, Guy Lorberbom, Idan Schwartz, and Tamir Hazan. 2021.
\newblock Latent space explanation by intervention.

\bibitem[{Geiger et~al.(2021)Geiger, Lu, Icard, and Potts}]{geiger_causal}
Atticus Geiger, Hanson Lu, Thomas~F Icard, and Christopher Potts. 2021.
\newblock \href {https://openreview.net/forum?id=RmuXDtjDhG} {Causal
  abstractions of neural networks}.
\newblock In \emph{Advances in Neural Information Processing Systems}.

\bibitem[{Geiger et~al.(2020)Geiger, Richardson, and Potts}]{geiger}
Atticus Geiger, Kyle Richardson, and Christopher Potts. 2020.
\newblock \href {https://doi.org/10.18653/v1/2020.blackboxnlp-1.16} {Neural
  natural language inference models partially embed theories of lexical
  entailment and negation}.
\newblock In \emph{Proceedings of the Third BlackboxNLP Workshop on Analyzing
  and Interpreting Neural Networks for NLP}, pages 163--173, Online.
  Association for Computational Linguistics.

\bibitem[{Giulianelli et~al.(2018)Giulianelli, Harding, Mohnert, Hupkes, and
  Zuidema}]{giulianelli_hood}
Mario Giulianelli, Jack Harding, Florian Mohnert, Dieuwke Hupkes, and Willem
  Zuidema. 2018.
\newblock \href {https://doi.org/10.18653/v1/W18-5426} {Under the hood: Using
  diagnostic classifiers to investigate and improve how language models track
  agreement information}.
\newblock In \emph{Proceedings of the 2018 {EMNLP} Workshop {B}lackbox{NLP}:
  Analyzing and Interpreting Neural Networks for {NLP}}, pages 240--248,
  Brussels, Belgium. Association for Computational Linguistics.

\bibitem[{Goyal et~al.(2019)Goyal, Feder, Shalit, and Kim}]{cace}
Yash Goyal, Amir Feder, Uri Shalit, and Been Kim. 2019.
\newblock Explaining classifiers with causal concept effect (cace).
\newblock \emph{arXiv preprint arXiv:1907.07165}.

\bibitem[{Hewitt and Liang(2019)}]{hewitt-liang}
John Hewitt and Percy Liang. 2019.
\newblock \href {https://doi.org/10.18653/v1/D19-1275} {Designing and
  interpreting probes with control tasks}.
\newblock In \emph{Proceedings of the 2019 Conference on Empirical Methods in
  Natural Language Processing and the 9th International Joint Conference on
  Natural Language Processing (EMNLP-IJCNLP)}, pages 2733--2743, Hong Kong,
  China. Association for Computational Linguistics.

\bibitem[{Kim et~al.(2018)Kim, Wattenberg, Gilmer, Cai, Wexler, Viegas
  et~al.}]{tcav}
Been Kim, Martin Wattenberg, Justin Gilmer, Carrie Cai, James Wexler, Fernanda
  Viegas, et~al. 2018.
\newblock Interpretability beyond feature attribution: Quantitative testing
  with concept activation vectors (tcav).
\newblock In \emph{International conference on machine learning}, pages
  2668--2677. PMLR.

\bibitem[{Liu et~al.(2019)Liu, Ott, Goyal, Du, Joshi, Chen, Levy, Lewis,
  Zettlemoyer, and Stoyanov}]{roberta}
Y.~Liu, Myle Ott, Naman Goyal, Jingfei Du, Mandar Joshi, Danqi Chen, Omer Levy,
  M.~Lewis, Luke Zettlemoyer, and Veselin Stoyanov. 2019.
\newblock Roberta: A robustly optimized bert pretraining approach.
\newblock \emph{ArXiv}, abs/1907.11692.

\bibitem[{MacCartney and Manning(2007)}]{maccartney-manning}
Bill MacCartney and Christopher~D. Manning. 2007.
\newblock \href {https://www.aclweb.org/anthology/W07-1431} {Natural logic for
  textul inference}.
\newblock In \emph{Proceedings of the {ACL}-{PASCAL} Workshop on Textual
  Entailment and Paraphrasing}, pages 193--200, Prague. Association for
  Computational Linguistics.

\bibitem[{Pimentel et~al.(2020)Pimentel, Valvoda, Hall~Maudslay, Zmigrod,
  Williams, and Cotterell}]{pimentel_itprobing}
Tiago Pimentel, Josef Valvoda, Rowan Hall~Maudslay, Ran Zmigrod, Adina
  Williams, and Ryan Cotterell. 2020.
\newblock \href {https://doi.org/10.18653/v1/2020.acl-main.420}
  {Information-theoretic probing for linguistic structure}.
\newblock In \emph{Proceedings of the 58th Annual Meeting of the Association
  for Computational Linguistics}, pages 4609--4622, Online. Association for
  Computational Linguistics.

\bibitem[{Ravfogel et~al.(2020)Ravfogel, Elazar, Gonen, Twiton, and
  Goldberg}]{ravfogel_nullspace}
Shauli Ravfogel, Yanai Elazar, Hila Gonen, Michael Twiton, and Yoav Goldberg.
  2020.
\newblock \href {https://www.aclweb.org/anthology/2020.acl-main.647/} {Null it
  out: Guarding protected attributes by iterative nullspace projection}.
\newblock In \emph{Proceedings of the 58th Annual Meeting of the Association
  for Computational Linguistics, {ACL} 2020, Online, July 5-10, 2020}, pages
  7237--7256. Association for Computational Linguistics.

\bibitem[{Ravichander et~al.(2021)Ravichander, Belinkov, and
  Hovy}]{ravichander_probing}
Abhilasha Ravichander, Yonatan Belinkov, and Eduard Hovy. 2021.
\newblock \href {https://doi.org/10.18653/v1/2021.eacl-main.295} {Probing the
  probing paradigm: Does probing accuracy entail task relevance?}
\newblock In \emph{Proceedings of the 16th Conference of the European Chapter
  of the Association for Computational Linguistics: Main Volume}, pages
  3363--3377, Online. Association for Computational Linguistics.

\bibitem[{Rozanova et~al.(2021{\natexlab{a}})Rozanova, Ferreira, Thayaparan,
  Valentino, and Freitas}]{rozanova_supporting}
Julia Rozanova, Deborah Ferreira, Mokanarangan Thayaparan, Marco Valentino, and
  André Freitas. 2021{\natexlab{a}}.
\newblock \href {http://arxiv.org/abs/2105.08008} {Supporting context
  monotonicity abstractions in neural nli models}.

\bibitem[{Rozanova et~al.(2021{\natexlab{b}})Rozanova, Ferreira, Valentino,
  Thayaparan, and Freitas}]{rozanova_decomposing}
Julia Rozanova, Deborah Ferreira, Marco Valentino, Mokanarangan Thayaparan, and
  Andr{\'{e}} Freitas. 2021{\natexlab{b}}.
\newblock \href {http://arxiv.org/abs/2112.08289} {Decomposing natural logic
  inferences in neural {NLI}}.
\newblock \emph{CoRR}, abs/2112.08289.

\bibitem[{Voita and Titov(2020)}]{voita-titov}
Elena Voita and Ivan Titov. 2020.
\newblock \href {https://doi.org/10.18653/v1/2020.emnlp-main.14}
  {Information-theoretic probing with minimum description length}.
\newblock In \emph{Proceedings of the 2020 Conference on Empirical Methods in
  Natural Language Processing (EMNLP)}, pages 183--196, Online. Association for
  Computational Linguistics.

\bibitem[{Williams et~al.(2018)Williams, Nangia, and Bowman}]{mnli}
Adina Williams, Nikita Nangia, and Samuel Bowman. 2018.
\newblock \href {http://aclweb.org/anthology/N18-1101} {A broad-coverage
  challenge corpus for sentence understanding through inference}.
\newblock In \emph{Proceedings of the 2018 Conference of the North American
  Chapter of the Association for Computational Linguistics: Human Language
  Technologies, Volume 1 (Long Papers)}, pages 1112--1122. Association for
  Computational Linguistics.

\bibitem[{Yanaka et~al.(2019)Yanaka, Mineshima, Bekki, Inui, Sekine,
  Abzianidze, and Bos}]{yanaka_help}
Hitomi Yanaka, Koji Mineshima, Daisuke Bekki, Kentaro Inui, Satoshi Sekine,
  Lasha Abzianidze, and Johan Bos. 2019.
\newblock \href {https://doi.org/10.18653/v1/S19-1027} {{HELP}: A dataset for
  identifying shortcomings of neural models in monotonicity reasoning}.
\newblock In \emph{Proceedings of the Eighth Joint Conference on Lexical and
  Computational Semantics (*{SEM} 2019)}, pages 250--255, Minneapolis,
  Minnesota. Association for Computational Linguistics.

\bibitem[{Zhu and Rudzicz(2020)}]{zhu_probing}
Zining Zhu and Frank Rudzicz. 2020.
\newblock \href {https://doi.org/10.18653/v1/2020.emnlp-main.744} {An
  information theoretic view on selecting linguistic probes}.
\newblock In \emph{Proceedings of the 2020 Conference on Empirical Methods in
  Natural Language Processing (EMNLP)}, pages 9251--9262, Online. Association
  for Computational Linguistics.

\end{thebibliography}
\bibliographystyle{acl_natbib}

\appendix
\section{Expanded Amnesic Intervention Results}
\begin{figure*}[h!]
\centering
\begin{subfigure}{.48\textwidth}
  \centering
  \includegraphics[width=\textwidth]{images/amnesic_ins_rel.png}
  \caption{Lexical Relation Probing Performance During Iterative Amnesic Intervention Process}
\end{subfigure}\hfill
\begin{subfigure}{.48\textwidth}
  \centering
  \includegraphics[width=\textwidth]{images/nli_xy_amnesic_ins_rel.png}
  \caption{Downstream Performance On NLI\_XY After Amnesic Intervention (Removing Lexical Relation Information)}
\end{subfigure}
\begin{subfigure}{.48\textwidth}
  \centering
  \includegraphics[width=\textwidth]{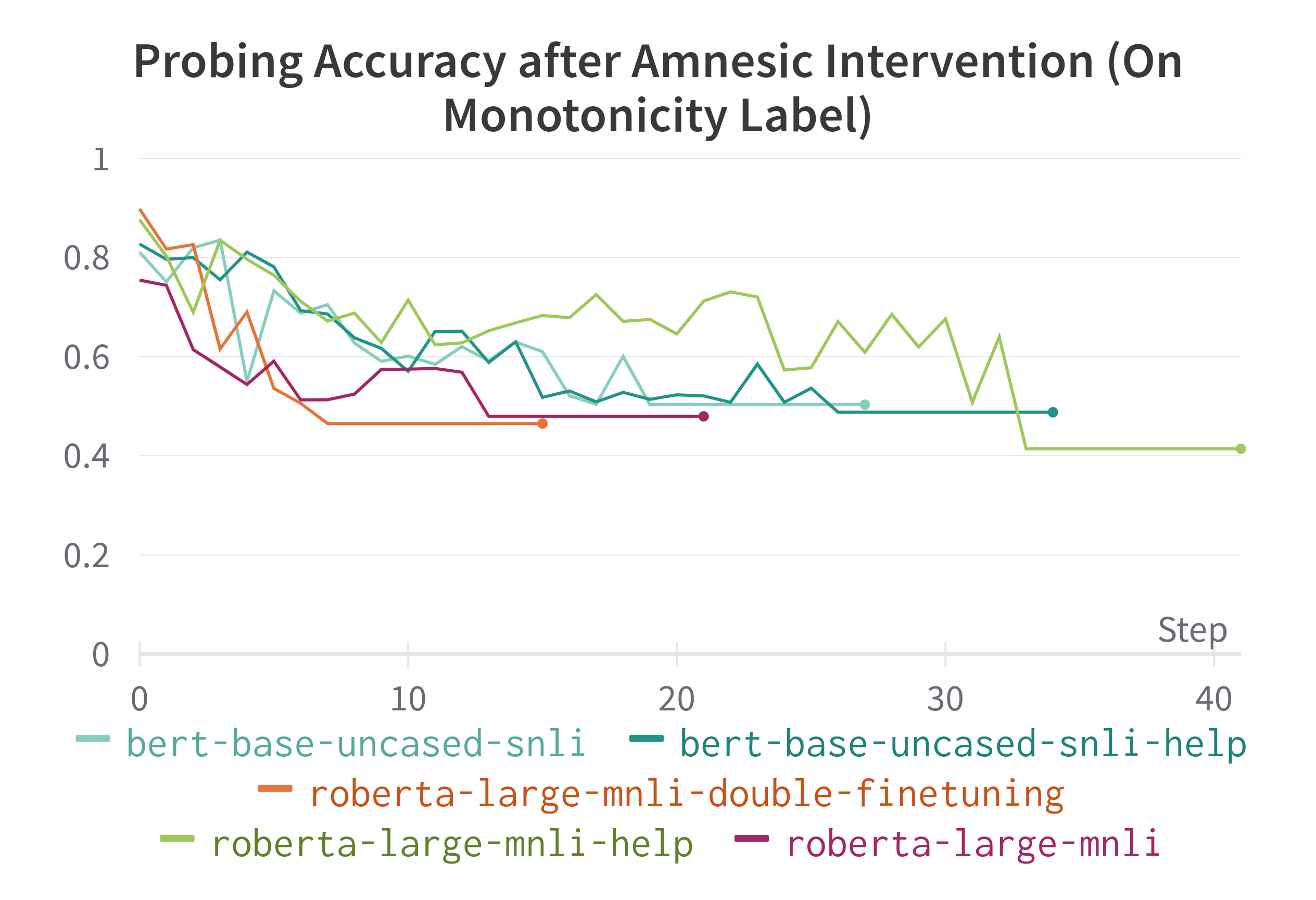}
  \caption{Context Monotonicity Probing Performance During Iterative Amnesic Intervention Process}
\end{subfigure}\hfill
\begin{subfigure}{.48\textwidth}
  \centering
  \includegraphics[width=\textwidth]{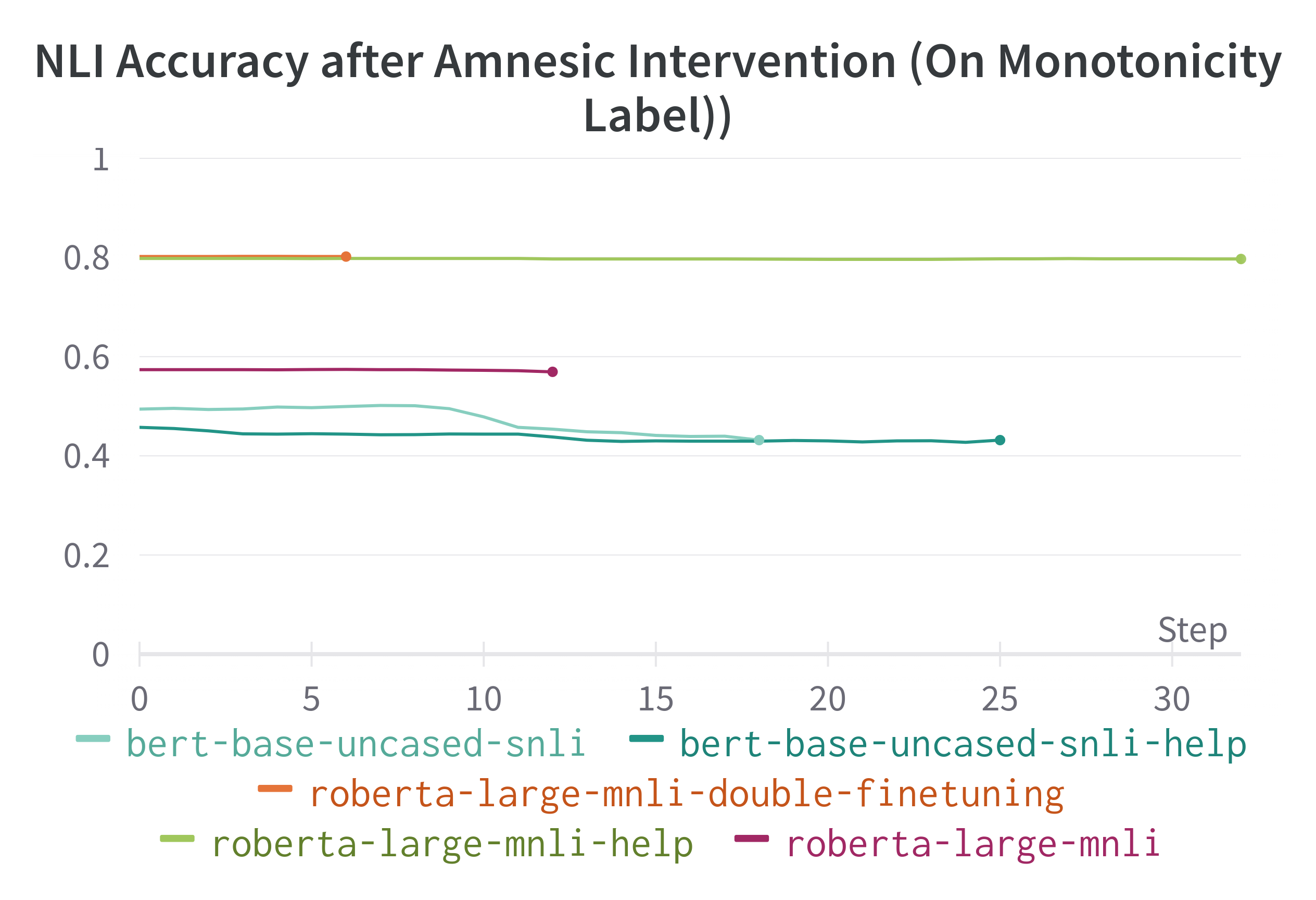}
  \caption{Downstream Performance On NLI\_XY After Amnesic Intervention (Removing Context Monotonicity Information)}
\end{subfigure}
\caption{Single Feature Amnesic Probing}
\label{fig:single_amnesic}
\end{figure*}

\begin{figure*}[ht!]
\begin{subfigure}{.48\textwidth}
  \centering
  \includegraphics[width=\textwidth]{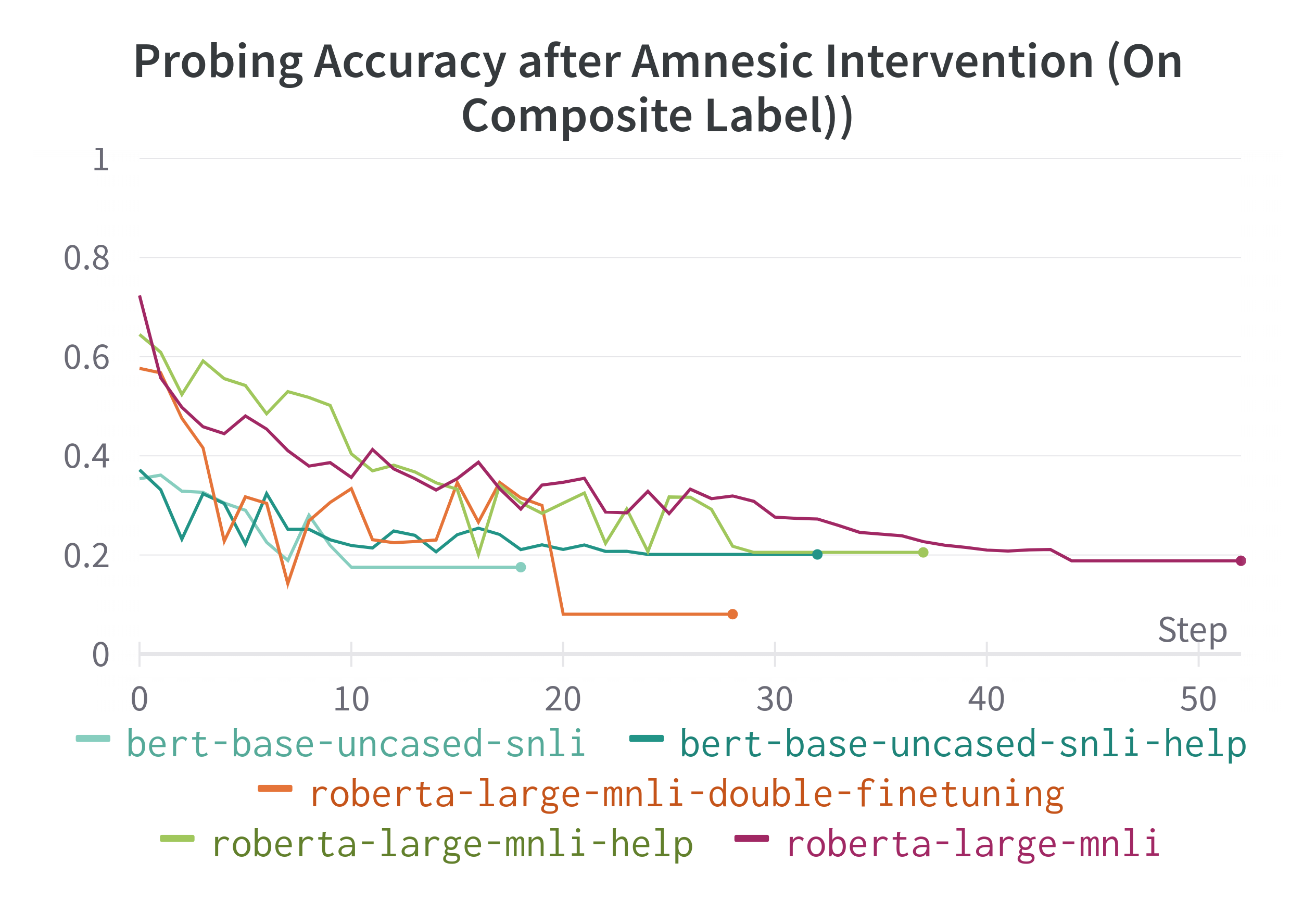}
  \caption{Probing Performance On NLI\_XY After Composite Label Amnesic Intervention}
\end{subfigure}\hfill%
\begin{subfigure}{.48\textwidth}
  \centering
  \includegraphics[width=\textwidth]{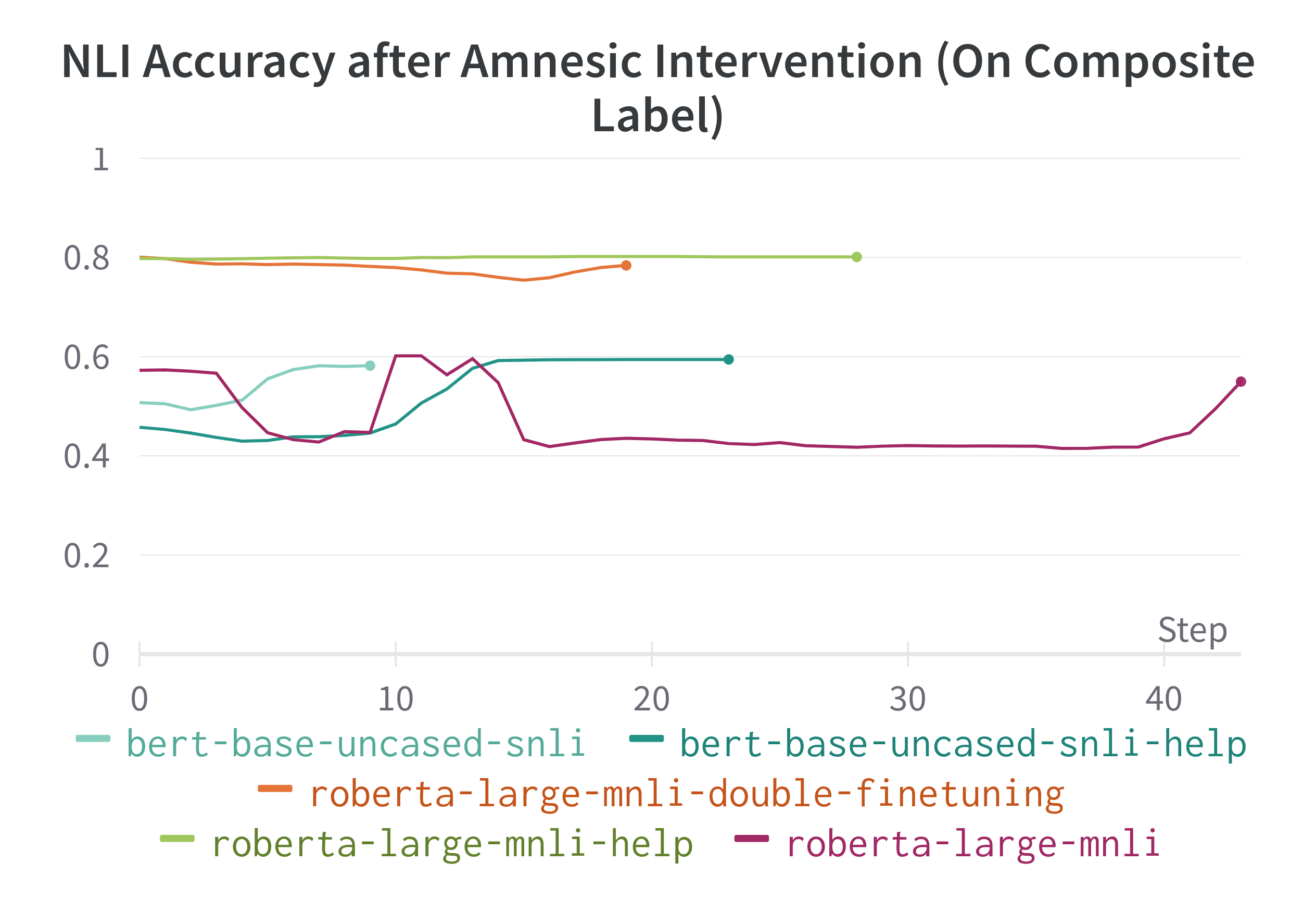}
  \caption{Downstream Performance On NLI\_XY After Composite Label Amnesic Intervention}
\end{subfigure}%
\caption{Composite Feature Label Amnesic Probing}
\label{fig:multi_amnesic}
\end{figure*}

\begin{figure*}[!htb]
\begin{subfigure}{.48\textwidth}
  \centering
  \includegraphics[width=\textwidth]{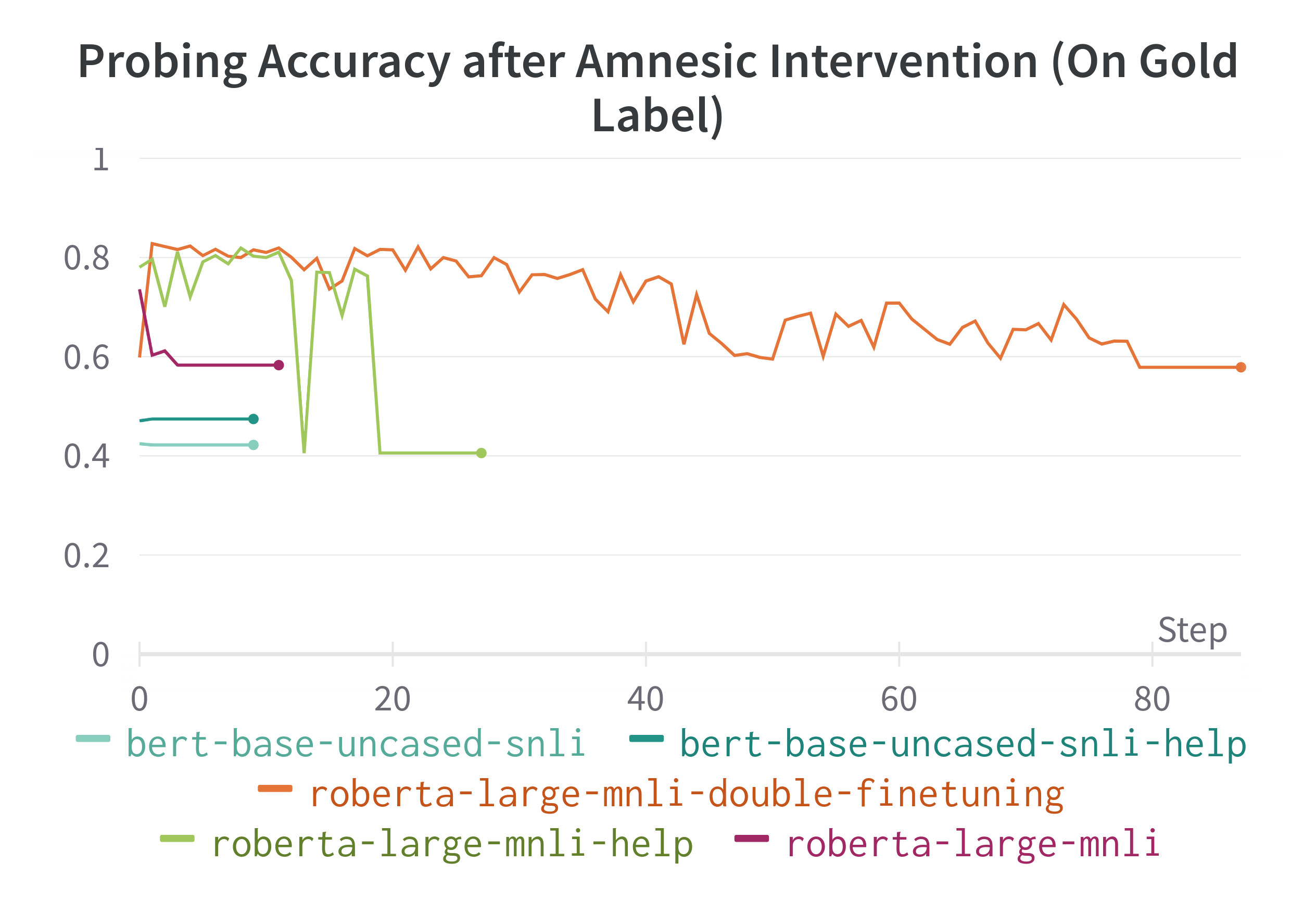}
  \caption{Probing performance On NLI\_XY after entailment label amnesic intervention.}
\end{subfigure}\hfill%
\begin{subfigure}{.48\textwidth}
  \centering
  \includegraphics[width=\textwidth]{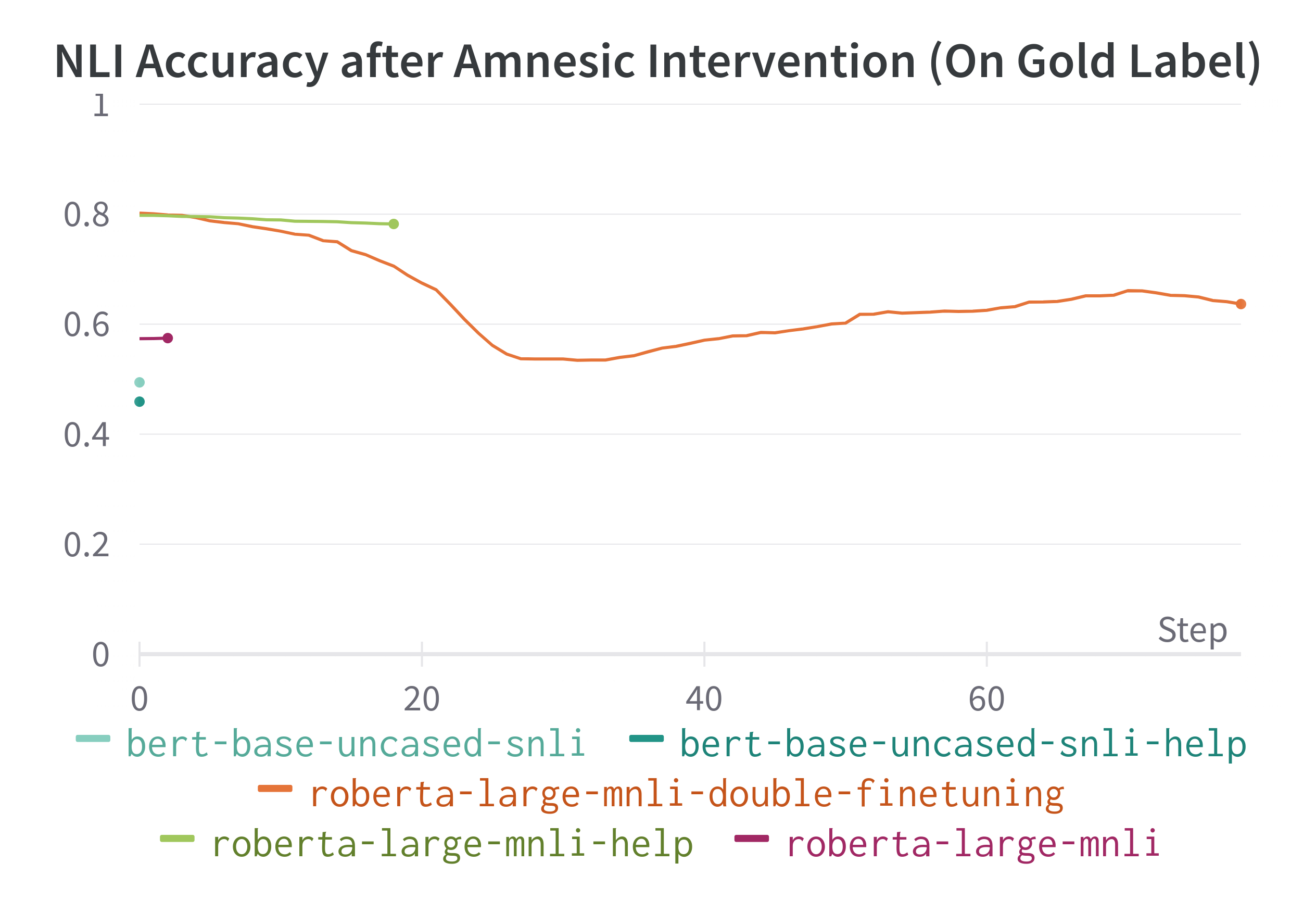}
  \caption{Downstream performance on NLI\_XY after entailment label amnesic intervention.}
\end{subfigure}%
\caption{Sanity Check: Entailment Gold Label Amnesic Probing}
\label{fig:multi_amnesic}
\end{figure*}

\end{document}